\documentclass[runningheads]{llncs}

\usepackage{eccv}

\usepackage{eccvabbrv}

\usepackage{graphicx}
\usepackage{booktabs}
\usepackage{multirow}

\usepackage[accsupp]{axessibility}  % Improves PDF readability for those with disabilities.

\usepackage{hyperref}

\usepackage{orcidlink}

\begin{document}

% ---------------------------------------------------------------
% TODO REVIEW: Replace with your title
\title{Weakly-Supervised 3D Hand Reconstruction with Knowledge Prior and Uncertainty Guidance} 

% TODO REVIEW: If the paper title is too long for the running head, you can set
% an abbreviated paper title here. If not, comment out.
\titlerunning{3D Hand Reconstruction with Knowledge and Uncertainty}

% TODO FINAL: Replace with your author list. 
% Include the authors' OCRID for the camera-ready version, if at all possible.
\author{Yufei Zhang\inst{1} \and
Jeffrey O. Kephart\inst{2} \and
Qiang Ji\inst{1}}

% TODO FINAL: Replace with an abbreviated list of authors.
\authorrunning{Y.~Zhang et al.}
% First names are abbreviated in the running head.
% If there are more than two authors, 'et al.' is used.

% TODO FINAL: Replace with your institution list.
\institute{$^1$~Rensselaer Polytechnic Institute, $^2$~IBM Research \\
\email{\{zhangy76, jiq\}@rpi.edu, kephart@us.ibm.com}}

\maketitle

\begin{abstract}
Fully-supervised monocular 3D hand reconstruction is often difficult because capturing the requisite 3D data entails deploying specialized equipment in a controlled environment. We introduce a weakly-supervised method that avoids such requirements by leveraging fundamental principles well-established in the understanding of the human hand's unique structure and functionality. Specifically, we systematically study hand knowledge from different sources, including biomechanics, functional anatomy, and physics. We effectively incorporate these valuable foundational insights into 3D hand reconstruction models through an appropriate set of differentiable training losses. This enables training solely with readily-obtainable 2D hand landmark annotations and eliminates the need for expensive 3D supervision. Moreover, we explicitly model the uncertainty that is inherent in image observations. We enhance the training process by exploiting a simple yet effective Negative Log-Likelihood (NLL) loss that incorporates uncertainty into the loss function. Through extensive experiments, we demonstrate that our method significantly outperforms state-of-the-art weakly-supervised methods. For example, our method achieves nearly a 21\% performance improvement on the widely adopted FreiHAND dataset.

\end{abstract}

\section{Introduction}
\label{sec:intro}

Reconstructing the 3D configuration of human hands has broad applications, especially for Virtual/Augmented Reality (VR/AR)~\cite{grubert2018effects,bai2020user} and Human-Computer Interaction (HCI)~\cite{sridhar2015investigating,parelli2020exploiting}. Traditional approaches rely on depth sensors~\cite{ballan2012motion,qian2014realtime,oberweger2016efficiently,yuan2017bighand2,moon2018v2v,yuan2018depth,garcia2018first,mueller2017real,oberweger2019generalized} or multi-camera setups~\cite{de2006regression,sridhar2014real,zhang20163d}. Due to their reliance on specialized equipment that is often expensive or unavailable, the practicality of such approaches is limited. We instead focus on monocular 3D hand reconstruction, reconstructing 3D hands from a single RGB image.

\begin{figure}[tb]
\begin{center}
\includegraphics[width=0.9\linewidth]{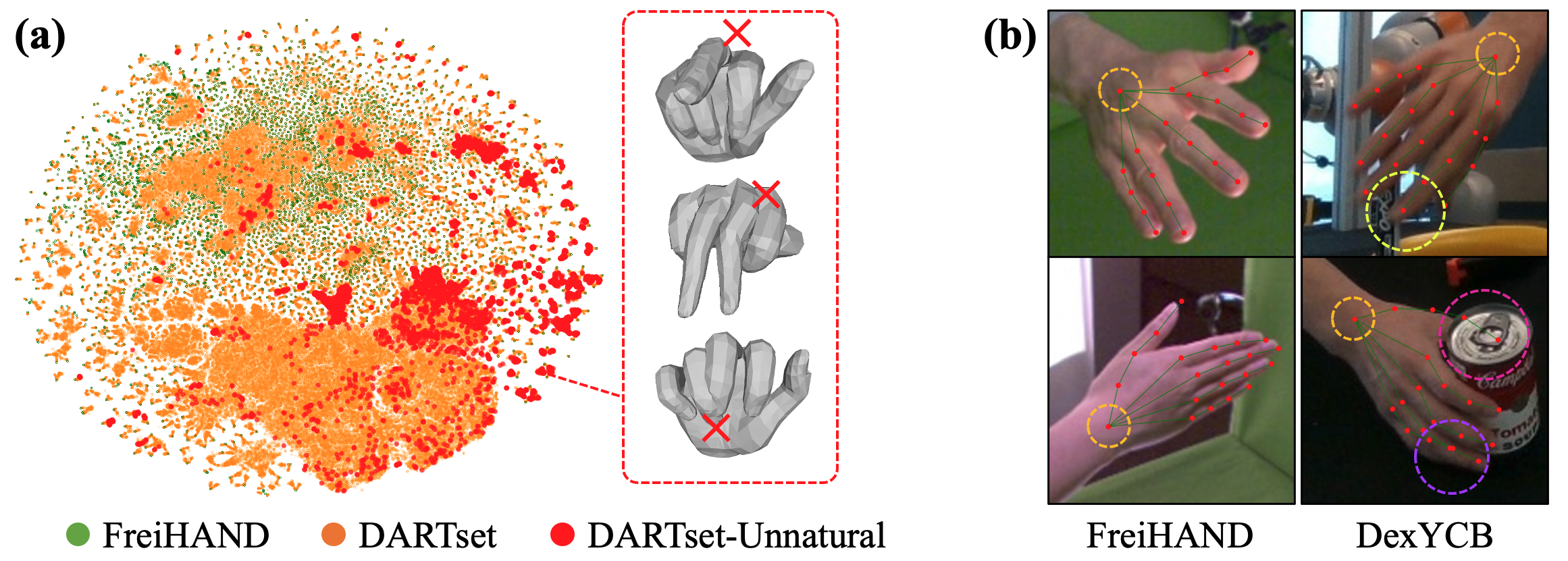}
\caption{\textbf{Motivation of Studying Hand knowledge and Modeling Uncertainty.} (a) T-SNE visualization~\cite{van2008visualizing} of hand poses from a real dataset (FreiHAND~\cite{zimmermann2019freihand}) and a synthetic dataset (DARTset~\cite{gao2022dart}). A large portion of synthetically generated hand poses can be unnatural (DARTset-Unnatural), such as the presence of invalid bending or penetration (marked by red crosses). (b) Images and 2D hand label from existing hand datasets. We mark regions with high uncertainty attributed to self-similarity (orange), motion blur (yellow), occlusion (pink), or poor image quality (purple).}
\label{fig:intro}
\end{center}
\end{figure}

Due to the lack of depth information in recovering 3D geometry from its 2D observation, monocular 3D hand reconstruction poses an ill-posed problem. Recent methods tackle this issue using deep learning models that predict 3D hand joint positions~\cite{zimmermann2017learning,spurr2018cross,spurr2018cross,yang2019aligning,yang2019disentangling,kim2021end,fan2021learning,meng20223d} or reconstruct a dense 3D hand mesh~\cite{zhou2020monocular,moon2020i2l,choi2020pose2mesh,hasson2020leveraging,lin2021end,spurr2021self,liu2021semi,moon2023bringing}. While these methods avoid the need for specialized equipment in constrained environments at inference time, they still rely upon it to obtain the 3D annotations required for {\em training} the deep models. The resulting limitations in the diversity and amount of data restrict the performance of these purely data-driven deep models. To address this challenge, some methods~\cite{mueller2018ganerated,gao2022dart,moon2023dataset,li2023renderih} leverage synthetically generated training images. The synthetic data are rich in quantity, but limited in the realism of the images and hand poses. As illustrated in Fig.~\ref{fig:intro}(a), many synthetically generated poses in DARTset appear unnatural due to a lack of systematic consideration of well-established principles of hand structure, functionality and movement during the data generation process. Additionally, approaches based on generating synthetic data still require some real 3D data for further model fine-tuning.

Other authors~\cite{boukhayma20193d, kulon2020weakly, spurr2020weakly, chen2021model} have exploited weakly-supervised learning, whereby the models are trained on real images with 2D hand landmark annotations. The advantage of such approaches is that 2D hand landmark labels are much more readily acquired in practice than 3D annotations.
Weakly-supervised 3D hand models are typically trained by minimizing two loss terms: (1) a prior term imposed on 3D hand prediction to encourage its realism under weak supervision, and (2) a data term measuring the consistency between the projection of 3D prediction and 2D image observations. When constructing the prior term, some methods~\cite{boukhayma20193d,baek2020weakly} learn the prior from data. However, there is no sufficiently homogeneous and dense data set that precisely captures realistic hand movement patterns, and moreover it is a significant challenge to acquire such data~\cite{herda2005hierarchical}. Other works~\cite{spurr2020weakly,chen2021model} attempt to derive the prior from hand literature, but they are often limited to a certain type of knowledge. Another issue exhibited in existing weakly-supervised approaches lies in their formulation of the data term. They overlook the uncertainty in image observations and employ standard regression losses, such as Mean Square Error (MSE). As shown in Fig.~\ref{fig:intro}(b), various types of image ambiguities may be present, posing a significant challenge to the reconstruction process. Failing to address such inherent uncertainty may lead to degraded model performance~\cite{kendall2017uncertainties}.

In this paper, we address the two issues prevalent in current weakly-supervised 3D hand reconstruction models by (1) systematically and effectively leveraging well-established knowledge about the human hand, and (2) explicitly modeling the uncertainty inherent in input images. Our method draws inspiration from KNOWN~\cite{zhang2023body}, which leverages body-specific knowledge and uncertainty for human body reconstruction. Here we adapt that approach to the hand. Specifically, we extract from a comprehensive study of literature on hand biomechanics, functional anatomy, and physics a useful body of hand knowledge. We encode it as a set of differentiable losses to enable training on images solely with 2D weak supervision. Moreover, we consider that the observation uncertainty varies at different hand joints for different input images. We model such heteroscedastic uncertainty by capturing the distribution of 2D hand landmark positions. We improve the training by exploiting a simple yet effective Negative Log-Likelihood (NLL) loss that automatically assigns weights to different 2D labels based on their captured uncertainty. Through extensive experiments, we demonstrate the effectiveness of the proposed method and its significant improvements over the existing weakly-supervised 3D hand reconstruction models.

In summary, our main contributions lie in: 
\begin{itemize}
    \item identifying valuable generic knowledge from a comprehensive study of hand literature, including hand biomechanics, functional anatomy, and physics;
    \item introducing a set of differentiable training losses to effectively integrate the identified knowledge into 3D hand reconstruction models;
    \item exploiting a simple yet effective NLL loss that incorporates the uncertainty in image observations to improve the training; and
    \item showing through extensive experiments that our method significantly outperforms existing methods under the challenging weakly-supervised setting.
\end{itemize}

\section{Related Work}
\label{sec:relatedwork}

In this section, we discuss recent advancements in monocular 3D hand reconstruction, considering fully-supervised and weakly-supervised settings.

\subsection{Fully-Supervised Approaches} Fully-supervised 3D hand reconstruction requires that 3D labels, such as ground truth 3D hand meshes, are sufficiently available. They focus on designing different model architectures for improved performance. One line of work follows a model-based reconstruction pipeline, wherein a 3D hand is represented by a deformable 3D hand model and reconstructed by estimating low-dimensional pose and shape parameters of the hand model~\cite{baek2019pushing,zhang2019end}. These model-based approaches can struggle to capture fine reconstruction details. Another line of work exploits a model-free reconstruction pipeline that directly predicts 3D hand mesh vertex positions~\cite{ge20193d, choi2020pose2mesh, moon2020i2l, lin2021mesh, lin2021end, chen2021camera, chen2022mobrecon, pavlakos2023reconstructing}. Such model-free approaches are typically data-hungry and less robust to occlusions and truncations. To address the issues inherent in both approaches, recent works~\cite{yu2023overcoming, jiang2023probabilistic} propose unifying the two pipelines into a single framework to enhance overall performance. Additionally, some models are specifically designed for handling cases like occlusion~\cite{park2022handoccnet}, hand-object interaction~\cite{hasson2019learning,tse2022collaborative,yang2022artiboost} or two-hand reconstruction~\cite{zhang2021interacting,li2022interacting,wang2023memahand,ren2023decoupled,yu2023acr,zuo2023reconstructing}. While such innovations improve estimation accuracy, none of them address the significant challenge of acquiring a sufficient amount of 3D data for fully-supervised learning.
%Other approaches have made promising stride by leveraging weak supervision, as discussed below.
%% JOK this sentence seems unnecessary because the reader already knows you are talking about weakly-supervised approaches below. If your meaning is that some fully-supervised approaches also make use of some weakly-supervised approaches, then it is fine to say that here but it needs to be clearer.

\subsection{Weakly-Supervised Approaches} Weak supervision approaches have made significant progress in enhancing the generalization and data efficiency of 3D reconstruction models~\cite{kanazawa2018end,zhang2023body}. In the context of 3D hand reconstruction, 2D hand landmark annotation proves to be a valuable form of weak supervision given its wide accessibility and the structural information it captures. Early works~\cite{boukhayma20193d,zhang2019end,chen2023mhentropy} relied on Principle Component Analysis (PCA) pose bases of the MANO hand model~\cite{romero2017embodied} and encouraged plausible 3D prediction by regularizing the prediction to be closer to the mean pose. Some works~\cite{zhang2019end,gao2022cyclehand} impose geometry constraints that assumed finger joints were located in the same plane during movement. Baek \textit{et al.}~\cite{baek2020weakly} propose capturing the complex 3D hand pose data distribution via Generative Adversarial Networks~\cite{goodfellow2020generative} and utilize the trained generative model as guidance for predicting realistic outputs. Instead of relying on data-driven priors or heuristic constraints, other works~\cite{mueller2018ganerated,kulon2020weakly,spurr2020weakly,chen2021model,ren2022end,tu2023consistent} impose joint rotation constraints with ranges retrieved from hand biomechanics literature and achieve improved performance. However, they overlook other sources of useful hand knowledge. Moreover, Tzionas \textit{et al.}~\cite{Tzionas:IJCV:2016} propose preventing invalid penetration in reconstructions by utilizing a non-penetration loss formulated over colliding mesh triangles. The proposed non-penetration loss only handles shallow penetration and cannot accommodate soft deformations that often occur in hand contact.

The contributions that differentiate our method from existing works are as follows. First, our study and utilization of generic hand knowledge is more comprehensive, and includes a novel inter-dependency derived from hand functional anatomy. Second, our encoding of knowledge is more effective. In particular, our formulation of the non-penetration loss effectively handles soft surface deformations by accurately pulling out deep inside vertices. Third, unlike existing works that neglect the heteroscedastic uncertainty in input images or limit their uncertainty modeling to hypothesis generation~\cite{chen2023mhentropy}, our method explicitly models the uncertainty and incorporates it into the training loss through a simple yet effective NLL loss, directly improving the training process. While this strategy has been studied in other applications~\cite{li2021human,zhang2023heteroscedastic,dwivedi2023poco,duan2024evidential}, we are the first to apply it to monocular 3D hand reconstruction.

\section{Method}
\label{sec:method}

\begin{figure}[t]
\begin{center}
\includegraphics[width=0.98\linewidth]{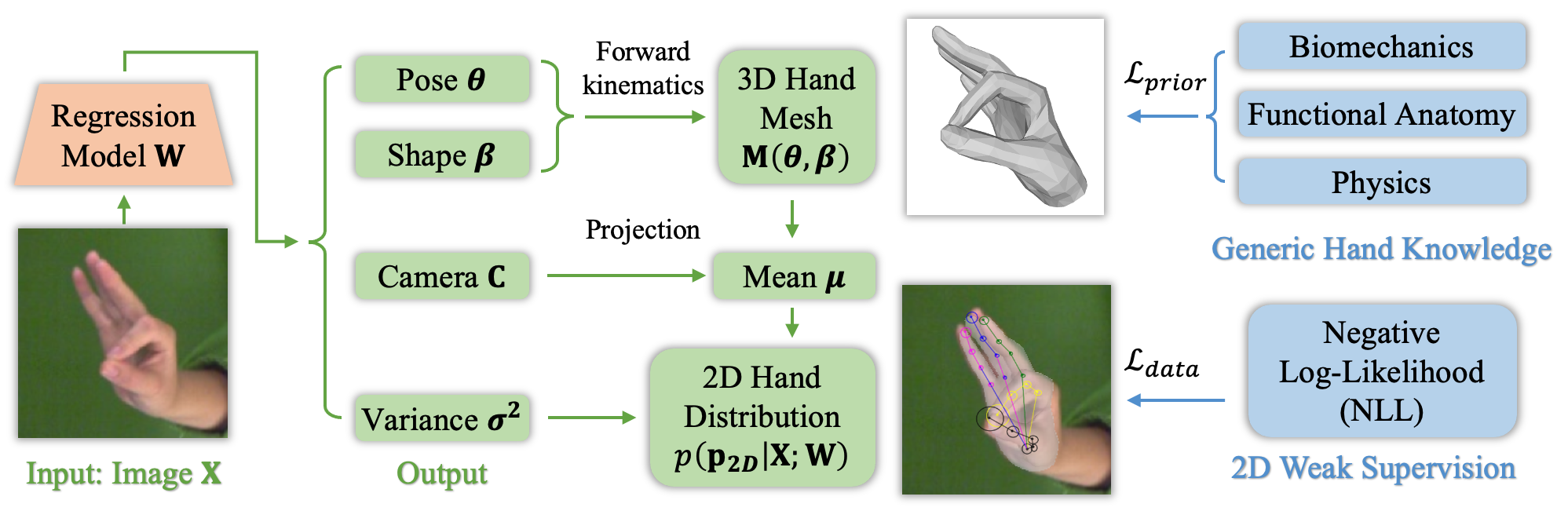}
\caption{\textbf{Overview of the proposed method.} Given a hand image, the regression model predicts the 3D hand pose and shape for recovering the 3D hand mesh through forward kinematics. The distribution of 2D hand landmark positions is specified via the projection of 3D hand and the predicted variance. The model is trained by incorporating generic hand knowledge and utilizing 2D hand landmark annotations.}
\label{fig:overview}
\end{center}
\end{figure}

Fig.~\ref{fig:overview} overviews our proposed method. We begin by introducing our 3D hand representation and camera projection model in Sec.~\ref{sec:preliminary}. Then, we systematically survey valuable hand knowledge and describe how we encode it as differentiable model training losses in Sec.~\ref{sec:knowledge}. We discuss the modeled distribution of 2D hand landmark positions and our formulation of the NLL loss in Sec.~\ref{sec:NLL}. Finally, we summarize the overall training loss for our model in Sec.~\ref{sec:overall}.

\subsection{Preliminaries}
\label{sec:preliminary}

\noindent\textbf{3D Hand Representation.} We employ MANO~\cite{romero2017embodied} to represent a 3D hand. MANO is a deformable 778-vertex 3D mesh model. It is parameterized by pose parameters $\boldsymbol{\theta}\in \mathbb{R}^{15\times3}$ that govern the rotation of 15 hand joints, and shape parameters $\boldsymbol{\beta}\in \mathbb{R}^{10}$ that represent the coefficients of PCA shape bases, capturing variations like hand length and width. Given $\boldsymbol{\theta}$ and $\boldsymbol{\beta}$, 3D mesh vertices $\mathbf{M}(\boldsymbol{\theta},\boldsymbol{\beta})\in \mathbb{R}^{778\times3}$ are obtained through forward kinematics. 3D hand joints $\mathbf{P}(\boldsymbol{\theta},\boldsymbol{\beta})\in \mathbb{R}^{J\times3}$ are a linear combination of the vertices as $\mathbf{P}(\boldsymbol{\theta},\boldsymbol{\beta})= \mathbf{H}\mathbf{M}(\boldsymbol{\theta},\boldsymbol{\beta})$, where $\mathbf{H}\in \mathbb{R}^{J\times778}$ is a joint regressor learned from data during the development of MANO, and $J=21$ indicates the number of modeled hand joints.

\noindent\textbf{Camera Projection Model.} Similar to existing practices, we estimate camera parameters $\mathbf{C}=[s,\mathbf{R},\mathbf{t}]$, where $s\in \mathbb{R}$, $\mathbf{R}\in\mathbb{R}^{3}$, and $\mathbf{t} \in \mathbb{R}^{2}$ denote the scale factor, camera rotation, and global translation, respectively. The projection of 3D hand joints is obtained as $\mathbf{p}_{2D}=Proj(\mathbf{P};\mathbf{C})$, where $Proj(\cdot)$ denotes the full-perspective projection function with a constant focal length, as in \cite{kanazawa2018end}.

\subsection{Study and Incorporation of Generic Hand Knowledge}
\label{sec:knowledge}

\begin{figure}[t]
\begin{center}
   \includegraphics[width=0.98\linewidth]{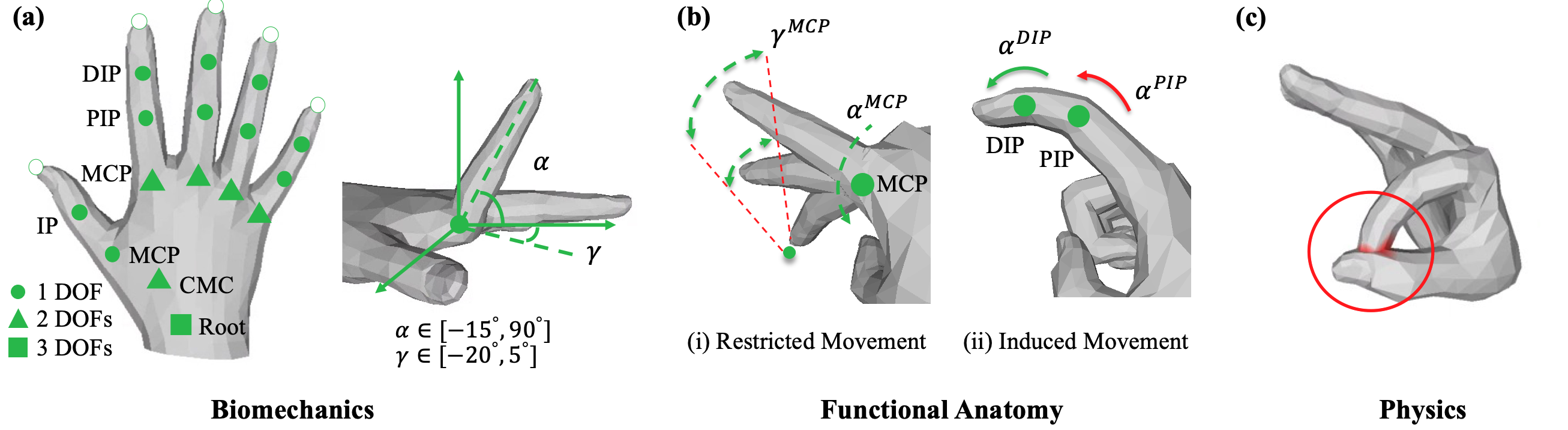}
   \caption{\textbf{Illustration of Generic Hand Knowledge.} (a) Different hand joints have different degrees of freedom (DOFs) and ranges of motion~\cite{hamill2006biomechanical}. (b) For the four fingers, (i) mutual restrictions exist between joint bending ($\alpha$) and splaying ($\gamma$) of the MCPs (metacarpophalangeal joints); (ii) the bending of the DIP (distal interphalangeal joint) induces bending in the PIP (proximal interphalangeal joint) due to tighter ligaments~\cite{schreuders2014functional}. (c) Different hand digits are prevented from penetrating into each other.}
    \label{fig:knowledge}
\end{center}
\end{figure}

Hand movement adheres to fundamental principles applicable across different subjects and gestures, serving as foundational insights for realistic 3D hands reconstruction. In this section, we systematically survey the generic hand knowledge from various sources, including hand biomechanics, functional anatomy, and physics. We introduce a set of differentiable losses over the 3D hand pose and shape parameters to integrate the knowledge into the reconstruction model.

\noindent \textbf{Hand Biomechanics} involves the quantitative study of hand movement mechanisms. There are 15 hand joints contributing to movement: a metacarpophalangeal joint (MCP), a proximal interphalangeal joint (PIP), and a distal interphalangeal joint (DIP) for each of the four fingers, and a carpometacarpal joint (CMC), an MCP, and an IP for the thumb. Each joint's movement can be described via three Euler angles corresponding to joint bending, splaying, and twisting, respectively. Hand biomechanics studies specify the DOFs and ranges of motion for each joint as illustrated in Fig.~\ref{fig:knowledge}(a). To impose these constraints, we introduce the following pose loss:
\begin{equation} 
\label{eq:bioloss}
    \mathcal{L}_{pose} = \sum_{j=1}^{15}(\max\{\boldsymbol{\theta}_{j}-\bar{\boldsymbol{\theta}}_{j,max},\bar{\boldsymbol{\theta}}_{j,min}-\boldsymbol{\theta}_{j},\mathbf{0}\})^2,
\end{equation}
where $\boldsymbol{\theta}_{j}$ represents the three Euler angles predicted for the $j^{th}$ joint. Their range, denoted by $(\bar{\boldsymbol{\theta}}_{j,min},\bar{\boldsymbol{\theta}}_{j,max})$, is obtained from literature~\cite{hamill2006biomechanical}, with the ranges set to zero for directions without degrees of freedom. Since the joint rotation coordinates used by MANO differ from those defined above, we adjust MANO's original coordinates by aligning its movement axes with the three Euler angles defined above. We also design the Euler angle rotation order for a joint based on their rotation range to avoid singularity, following \cite{zhang2023body,zhang2024incorporating,Zhang_2024_CVPR}.

\noindent \textbf{Hand Functional Anatomy} investigates how the hand's anatomical structure influences its movement. In contrast to hand biomechanics, which delineates the range of motion for each individual joint, the study of functional anatomy stipulates essential inter-joint dependencies during hand movement. As illustrated in Fig. \ref{fig:knowledge}(b), there are two types of dependencies: (i) the bending of the MCP restricts its splaying, following a linear relationship that peaks at the maximum bending angle~\cite{schultz1987metacarpophalangeal}; and (ii) the bending of the DIP induces bending of the PIP within the same finger~\cite{schreuders2014functional}. These inter-dependencies highlight that the range of motion of the hand joints can be dynamic and dependent on each other. Specifically, denote the predicted bending and splaying angles of an MCP joint $j$ as $\alpha_j^{MCP}$ and $\gamma_j^{MCP}$, respectively. Based on the Type-(i) dependency, their range of motion should be updated as:
\begin{equation}
\label{eq:functionanatomy1}
\begin{split}
    \hat{\gamma}_{j,min}^{MCP}&=\bar{\gamma}_{j,min}^{MCP}(1-\frac{\alpha_j^{MCP}}{\bar{\alpha}_{j,min}^{MCP}}), \ \text{if} \  \bar{\alpha}_{j,min}^{MCP}<\alpha_j^{MCP}<0, \\
    \hat{\gamma}_{j,max}^{MCP}&=\bar{\gamma}_{j,max}^{MCP}(1-\frac{\alpha_j^{MCP}}{\bar{\alpha}_{j,max}^{MCP}}), \ \text{if} \ 0<\alpha_j^{MCP}<\bar{\alpha}_{j,max}^{MCP}, \\
\end{split}
\end{equation}
where $(\bar{\gamma}_{j,min}^{MCP},\bar{\gamma}_{j,max}^{MCP})$ and $(\bar{\alpha}_{j,min}^{MCP},\bar{\alpha}_{j,max}^{MCP})$ are the ranges based on hand biomechanics, while $(\hat{\gamma}_{j,min}^{MCP},\hat{\gamma}_{j,max}^{MCP})$ denotes the refined range. As shown, the range of $\gamma_j^{MCP}$ becomes very limited as $\alpha_{j}^{MCP}$ approaches extreme angles. Similarly, the range of motion for $\alpha_j^{MCP}$ needs to be further constrained based on the value of $\gamma_j^{MCP}$. Moreover, denote the predicted bending of the PIP and DIP of finger $k$ as $\alpha_k^{PIP}$ and $\alpha_k^{DIP}$, respectively. According to the Type-(ii) dependency, the lower bound of $\alpha_k^{PIP}$ should be refined as:
\begin{equation}
\label{eq:functionanatomy2}
    \hat{\alpha}_{k,min}^{PIP} = 0, \  \text{if} \ \alpha_k^{DIP}>0,
\end{equation}
where $\alpha_k^{PIP}$ is encouraged to be greater than zero given a flexion DIP. In summary, the two types of dependencies refine the ranges ($\bar{\boldsymbol{\theta}}_{min}$, $\bar{\boldsymbol{\theta}}_{max}$) provided by the hand biomechanics to ($\hat{\boldsymbol{\theta}}_{min}$, $\hat{\boldsymbol{\theta}}_{max}$) based on the current hand pose prediction $\boldsymbol{\theta}$. To integrate this valuable anatomical knowledge into the 3D reconstruction model, we dynamically update the joint rotation ranges following Eq.~\ref{eq:functionanatomy1} and Eq.~\ref{eq:functionanatomy2}, and utilize the refined ranges to calculate the pose loss in Eq.~\ref{eq:bioloss}.

\noindent \textbf{Hand Physics} studies assert various principles governing the physical interactions of the human hand. As our model reconstructs a single 3D hand from a single image, we mainly consider static physics, particularly the principle of non-penetration, according to which different hand parts cannot penetrate into each other. Fig.~\ref{fig:knowledge}(c) illustrates a failure case. To integrate the non-penetration principle into the 3D reconstruction model, we first identify a set $\mathbf{M}$ comprising vertices located inside the mesh through the generalized winding number~\cite{jacobson2013robust,muller2021self}. For each vertex $\mathbf{v}\in \mathbf{M}$, we then apply the following non-penetration loss:
\begin{equation}
    \mathcal{L}_{non-penetration} = \sum_{\mathbf{v}\in \mathbf{M}}\max\{d(\mathbf{v})-d_{tol},0\},
\end{equation}
where $d(\mathbf{v})$ denotes the minimum distance from vertex $\mathbf{v}$ to another vertex that is not a neighbor of $\mathbf{v}$ (where the geodesic distance exceeds the average length of phalanges, e.g., 2cm). In other words, $d(\mathbf{v})$ represents the minimum distance from vertex $\mathbf{v}$ to 3D hand surface. Meanwhile, recognizing MANO's limitation in modeling soft surface deformations during contact, we introduce a tolerance distance $d_{tol}$ to accommodate shallow penetrations. Unlike existing methods~\cite{Tzionas:IJCV:2016} that formulate the loss based on collision triangles and only deter shallow penetrations, our proposed loss is applied to vertices with distances to the surface exceeding $d_{tol}$, effectively pulling out the those deeply embedded vertices.

\noindent \textbf{Overall Knowledge-Encoded Prior.} Incorporating the knowledge discussed above ensures natural 3D hand pose predictions. Similar to existing methods~\cite{kulon2020weakly,chen2021model}, we apply a shape regularization $\mathcal{L}_{shape} = \|\boldsymbol{\beta}\|_2$ to promote plausible hand shape predictions. Assembling all the losses together, we obtain the overall prior:
\begin{equation}
\label{eq:prior}
    \mathcal{L}_{prior} = \lambda_1\mathcal{L}_{pose}+\lambda_2\mathcal{L}_{non-penetration}+\lambda_3\mathcal{L}_{shape}
\end{equation}
It is worth noting that the prior term in Eq.~\ref{eq:prior} is derived from generic hand knowledge, which is applicable to all subjects and gestures. Notably, its formulation does not require any 3D data and is independent of any specific dataset.

\subsection{Training with Negative Log-Likelihood}
\label{sec:NLL}

To further ensure that predictions are consistent with the image observations, we utilize 2D hand landmark annotations. The input images can often exhibit challenges, such as occlusion or low image quality, that result in inherently ambiguous 2D hand positions or high uncertainty in the 3D reconstruction. Unlike existing methods that overlook this inherent uncertainty and train on 2D hand labels using standard regression loss, we explicitly model the uncertainty and incorporate it into the loss function to enhance model performance. Specifically, we model the uncertainty by capturing the distribution of 2D hand landmark positions. As different 2D hand landmarks exhibit different appearance features that vary across input images, we model each joint independently and capture input-dependent uncertainty. We assume the distribution of 2D hand landmark positions $\mathbf{p}_{2D}$ of an image $\mathbf{X}$ as:
\begin{equation}
\label{eq:distribution}
p(\mathbf{p}_{2D}|\mathbf{X};\mathbf{W}) = \prod_i\frac{1}{\sqrt{2\pi}\boldsymbol{\sigma}_i} \exp\left(-\frac{(\mathbf{p}_{2D,i} - \boldsymbol{\mu}_i)^2}{2\boldsymbol{\sigma}_i^2}\right),
\end{equation}
where $i$ is the image location index of hand joints. The adoption of Gaussian distributions is based on their wide utility in modeling observation noise~\cite{kendall2017uncertainties}. $\boldsymbol{\mu}$ represents the mean of the Gaussian distributions computed through the projection of 3D hand joint positions $\mathbf{P}$ using the camera parameters $\mathbf{C}$, while the variance $\boldsymbol{\sigma}^2$ are directly predicted by the regression model with parameters $\mathbf{W}$. 

The modeled distribution $p(\mathbf{p}_{2D}|\mathbf{X};\mathbf{W})$ specifies the probability of the ground truth appearing at position $\mathbf{p}_{2D}$. The labeled position $\bar{\mathbf{p}}_{2D}$ can be viewed as an observed data sample. We can thus train the model through Maximum Likelihood Estimation. It minimizes the Negative Log-Likelihood (NLL) to construct the data term to ensure 3D-2D consistency as:
\begin{equation}
\label{eq:NNL}
\begin{split}
    \mathcal{L}_{data}&=-\log{p(\mathbf{p}_{2D}=\bar{\mathbf{p}}_{2D}|\mathbf{X};\mathbf{W})} \\
 &\propto \sum_i{\left(\log{\boldsymbol{\sigma}_i} + \frac{(\bar{\mathbf{p}}_{2D,i} - \boldsymbol{\mu}_i)^2}{2\boldsymbol{\sigma}_i^2}\right)}.
\end{split}
\end{equation}
Note that the variance $\boldsymbol{\sigma}^2$ in Eq.~\ref{eq:NNL} depends on the individual hand joint $i$. Omitting the variance estimation or treating it as a constant would be equivalent to using the standard MSE loss, which is agnostic to uncertainty and assigns weights to all samples uniformly. In contrast, our method assigns reduced weights to images and joints with high uncertainty in a principled fashion, thereby producing a more robust model with improved performance.

\subsection{Total Training Loss}
\label{sec:overall}

By combining the prior term in Eq.~\ref{eq:prior} and the data term in Eq.~\ref{eq:NNL},  we obtain the total loss for training the regression model as: 
\begin{equation}
    \mathcal{L}=\mathcal{L}_{prior}+\mathcal{L}_{data}.
\end{equation}
During testing, 3D hands can be directly reconstructed through the hand pose and shape parameters estimated by the regression model.

\section{Experiment}
\label{sec:experiment}

We briefly introduce our data sets, evaluation metrics, and implementation details in Sec.~\ref{sec:data}. Then, in Sec.~\ref{sec:ablation}, we discuss an ablation study that demonstrates the effectiveness of incorporating various sources of generic hand knowledge and training with the Negative Log-Likelihood (NLL). Finally, in Sec.~\ref{sec:sota}, we assess the improved performance of our method in comparison to existing weakly-supervised State-of-the-Art (SOTA) approaches.

\subsection{Datasets, Metrics, and Implementation Details}
\label{sec:data}

\noindent\textbf{Datasets.} We employ three widely adopted datasets: FreiHAND~\cite{zimmermann2019freihand}, DexYCB~\cite{chao2021dexycb}, and HO3Dv3~\cite{hampali2021ho}, all of which have been captured by multi-view data collection systems. FreiHAND features a diverse range of daily hand poses. DexYCB and HO3Dv3 contain hand-object interaction images, some of which are significantly occluded. We follow the established training and testing splits to facilitate comparison with other methods.

\noindent\textbf{Evaluation Metrics.} Like existing methods~\cite{chen2021model,jiang2023probabilistic}, we compute the average Euclidean distance between the predicted and the ground truth 3D hand joint and vertex positions after procrustes alignment (E$_{J}$/E$_{V}$). The evaluation on HO3Dv3 is obtained through the online submission system. It further includes AUC$_{J}$/AUC$_{V}$, the area under the percentage of correct keypoint (PCK) curves with thresholds between 0mm and 50mm. %, and F$_{V}$-5/F$_{V}$-15, the F-score of vertices with thresholds of 5/15mm. 
Additionally, we compute penetration rate (PR), the percentage of reconstructions exhibiting penetration with a depth greater than $d_{tol}$, to assess the physical plausibility of 3D reconstructions.

%Additionally, we report the area under the curve (AUC) of the Percentage of Correct Keypoints (PCK) metric, where the thresholds for considering a 3D mesh vertex prediction as correct are set between 0mm to 50mm. Evaluation HO3Dv3 is obtained through the online submission system \url{https://codalab.lisn.upsaclay.fr/competitions/4393}.

\noindent\textbf{Implementation.} We implemented our framework using PyTorch. The regression model consists of a ResNet-50 model~\cite{he2016identity} %pretrained on ImageNet~\cite{deng2009imagenet} 
to extract image features and an iterative error feedback regression model~\cite{carreira2016human} to predict the unknown parameters from the extracted features. The hand images are scaled to $224 \times 224$ while preserving the aspect ratio. The training images are augmented with random scaling and flipping. The training batch size and epochs is 64 and 200, respectively. Following the training strategy in \cite{zhang2023body}, we initially employ the MSE for the data term and then utilize the NLL for faster convergence. We use the Adam optimizer~\cite{kingma2014adam} with a learning rate of $10^{-5}$ and weight decay of $10^{-4}$. The hyper-parameters are set to $d_{tol}=6mm$, $\lambda_1=20000$, $\lambda_2=20000$, and $\lambda_3=10$.

\subsection{Ablation Study}
\label{sec:ablation}

Table~\ref{tab:ablaknowledge} summarizes the impact of a) incorporating hand knowledge and b) training with the NLL loss. We provide a detailed analysis of these results below.

\begin{figure}[t!]
\begin{center}
\includegraphics[width=0.75\linewidth]{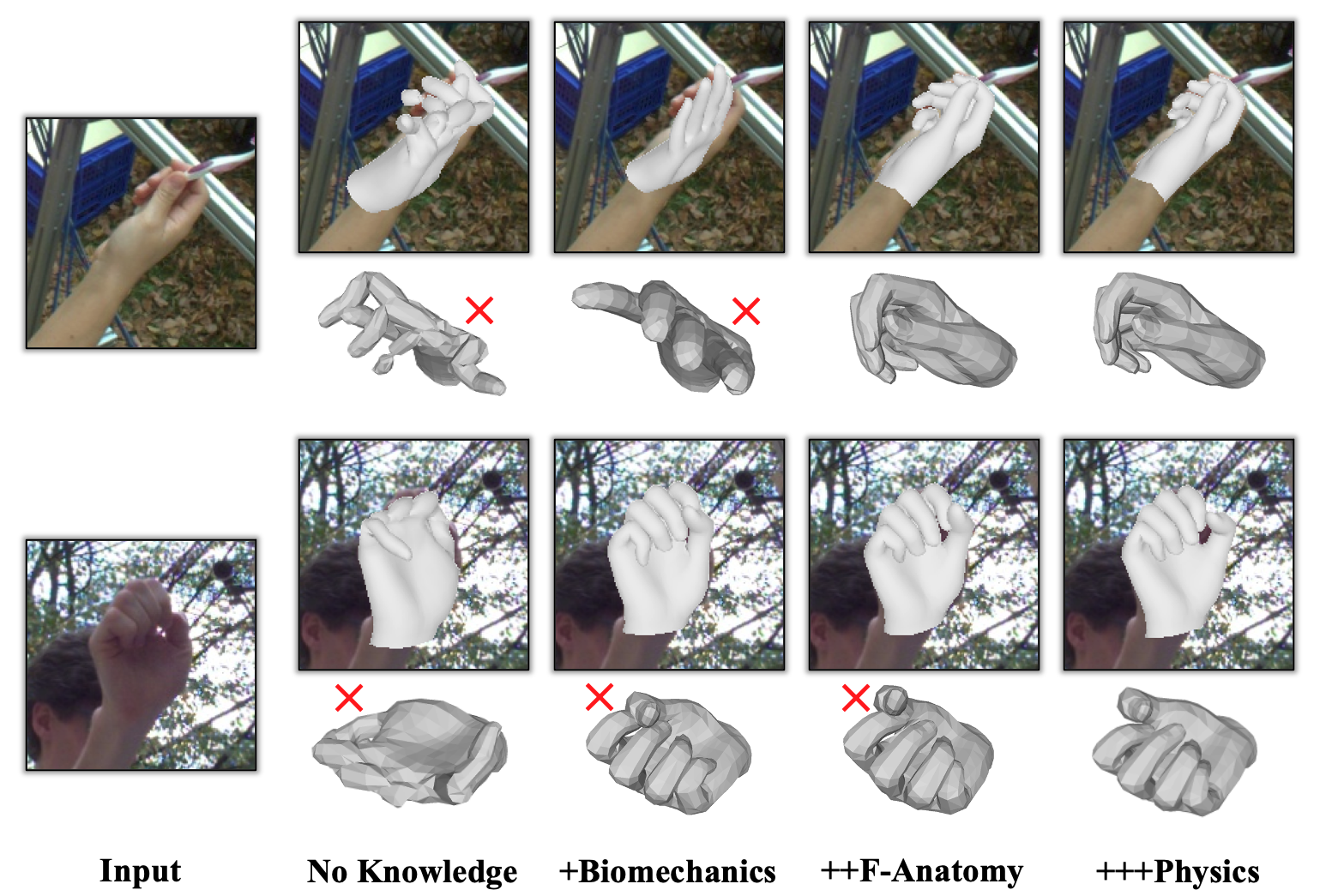}
\caption{\textbf{Qualitative Evaluation of Incorporating Hand Knowledge.} The images are from FreiHAND's test set. For each example, we present the rendered 3D hand overlaid on the input image, along with the reconstructed 3D hand viewed from a different angle. The results from left to right are obtained by incorporating the additional knowledge specified at the bottom. ``F-Anatomy'' denotes ``Functional Anatomy''. Reconstructions with notable errors are marked by red crosses.}
\label{fig:ablaknowledge}
\end{center}
\end{figure}

\noindent\textbf{Incorporating Generic Hand Knowledge.} To provide valuable insights about the effectiveness of leveraging different sources of hand knowledge, we supplement Table~\ref{tab:ablaknowledge} with a qualitative evaluation in Fig.~\ref{fig:ablaknowledge}. When not integrating any hand knowledge, the model is trained using 2D hand landmark annotations with a prior term that only includes the shape regularization. This model produces large reconstruction errors (Table~\ref{tab:ablaknowledge}, row1). As illustrated in Fig.~\ref{fig:ablaknowledge} (``No Knowledge''), the reconstructions can align with the image observations, but the predicted 3D hand poses are fairly unrealistic. The infeasible twisted fingers significantly violate the joint range of motion specified by hand biomechanics. This issue is addressed by introducing hand biomechanics into the training, resulting in a significant model performance boost (Table~\ref{tab:ablaknowledge}, row2 over row1). For example, E$_{J}$ is improved from 22.4mm to 10.9mm. Meanwhile, the estimated 3D hand poses become more plausible as shown in Fig.~\ref{fig:ablaknowledge} (``+Biomechanics''). Nonetheless, poor reconstructions can still occur due to the inherent depth ambiguity. Specifically, the relative depth of hand joints can be incorrect. Mitigating this issue requires the further incorporation of the functional anatomy knowledge (Fig.~\ref{fig:ablaknowledge}, ``++F-Anatomy''), leading to a reduction of the reconstruction errors from 10.9mm to 9.6mm for E$_{J}$ and from 11.4mm to 10.0mm for E$_{V}$ (Table~\ref{tab:ablaknowledge}, row3 over row2). The functional anatomy captures inter-joint dependency, alleviating the depth ambiguity by introducing additional constraints on the 3D reconstruction space. Furthermore. avoiding invalid penetrations in the reconstructions requires adding the proposed non-penetration loss (Fig.~\ref{fig:ablaknowledge}, bottom example). Incorporating this physics knowledge effectively reduces the percentage of reconstructions with invalid penetration from 11.4\% to 1.9\%. To a lesser degree, but still significantly, it also improves the other reconstruction accuracy metrics. In summary, by incorporating hand knowledge gleaned from literature into our proposed training loss functions, we are able to generate accurate 3D hand reconstruction models based solely on 2D weak supervision.

\begin{table}[tb]
\begin{center}
\caption{\textbf{Quantitative Evaluation of Incorporating Hand Knowledge and Training into the NLL Loss.} The evaluation is on FreiHAND. Without incorporating any hand knowledge, the model is trained by utilizing 2D hand landmark annotations and the shape regularization. ``F-Anatomy'' denotes ``Functional Anatomy''. The units of E$_{J}$ and E$_{V}$ are in mm, while PR is in percentage.}
\label{tab:ablaknowledge}
\begin{tabular}{ccc  ccc  ccc }
\toprule
\multicolumn{3}{c}{Hand Knowledge} & & Weak Supervision & & \multicolumn{3}{c}{Reconstruction} \\
\cmidrule(lr){1-3} \cmidrule(lr){4-6} \cmidrule(lr){7-9}
 Biomechanics & F-Anatomy & Physics & & NLL &  & E$_{J}$$\downarrow$ & E$_{V}$$\downarrow$ & PR$\downarrow$\\
\midrule
  &  & & && & 22.4 & 24.8 & 38.8 \\ 
$\checkmark$ & & & && & 10.9 & 11.4 & 11.4 \\
$\checkmark$ &  $\checkmark$ & & && & 9.6 & 10.0 & 11.4 \\ 
$\checkmark$ &  $\checkmark$ & $\checkmark$ & && & 9.4 & 
9.8 & 1.9 \\ 
\midrule
$\checkmark$ &  $\checkmark$ & $\checkmark$ & & $\checkmark$ & & \textbf{8.5} & 
\textbf{8.9} & \textbf{1.3} \\ 
\bottomrule
\end{tabular}
\end{center}
\end{table}

\begin{figure}[tb]
\begin{center}
\includegraphics[width=0.99\linewidth]{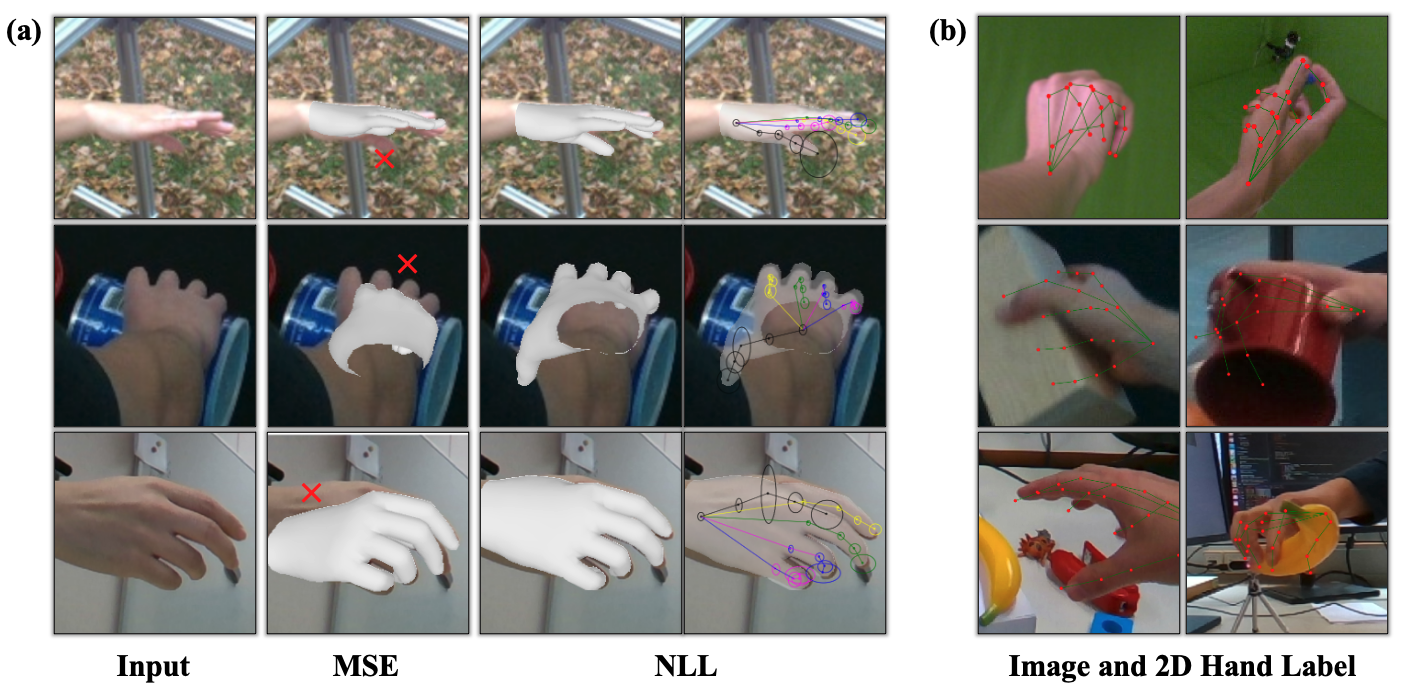}
\caption{\textbf{Qualitative Evaluation of Training with the NLL.} (a) Evaluation of the models trained without and with the NLL. Notable errors are marked by red crosses. Colors indicate finger identity: thumb (black), index (yellow), middle (green), ring (blue), and pinky (magenta). The width and height of the ellipse at each joint represent the magnitude of the estimated variance along the horizontal and vertical directions, respectively. (b) Training images with high uncertainty, captured by large estimated variances (averaged over all joints). The images in (a) and (b) are from FreiHAND (top), DexYCB (middle), and HO3D (bottom).}
\label{fig:ablanll}
\end{center}
\end{figure}

\noindent\textbf{Training with Negative Log-likelihood.} As discussed in Sec.~\ref{sec:NLL}, the NLL loss takes into account the increased reconstruction uncertainty of images containing occlusions or other degradations at the granularity of individual hand joints. Comparing rows 5 and 4 of Table~\ref{tab:ablaknowledge} we see that the 3D joint position error E$_{J}$ is reduced from 9.4mm to 8.5mm and the 3D mesh reconstruction error E$_{V}$ is decreased from 9.8mm to 8.9mm. In Fig.~\ref{fig:ablanll}(a), we provide qualitative comparison between the models trained without and with the NLL loss. When not utilizing the NLL, the model is trained using the MSE loss. As shown, such models can be adversely affected by low image quality, image occlusion, and image truncation occurring at various hand joints, resulting in poor 3D hand reconstructions. In contrast, the model trained with the NLL is more robust to these situations. For example, the alignment to the input image is significantly improved compared to training with the MSE even when the hand is heavily occluded (Fig.~\ref{fig:ablanll}(a), middle example). Furthermore, the model trained with the NLL captures the distribution of 2D hand positions. As shown in Fig.~\ref{fig:ablanll}(a) (column4), the hand joints in low-quality or occluded regions are captured by high variance estimates (visualized by large ellipses). In Fig.~\ref{fig:ablanll}(b), we present training images with large estimated variances. As shown, the images with excessive occlusion, truncation, and ambiguity appearance exhibit large variance estimations. During training, the utilization of the NLL effectively enhances the final model performance by incorporating the uncertainty into the training loss function.

\subsection{Comparison with State-of-the-Art}
\label{sec:sota}

\begin{table*}[tb]
\begin{center}
\caption{\textbf{Comparison with Weakly-Supervised SOTA Methods.} The evaluation of other methods are obtained from published papers. * denotes the methods using 3D annotations in synthetic or real 3D hand datasets during training. For the evaluation on DexYCB, we report the 3D mesh reconstruction error with the root translation alignment E$_{RV}$ to compare with others. The units of E$_{J}$, E$_{V}$, and E$_{RV}$ are in mm.}
\label{tab:sota}
\begin{tabular}{l cc cc cccc  }
\toprule
\multirow[b]{2}{*}{Method} & \multicolumn{2}{c}{FreiHAND} &  \multicolumn{2}{c}{DexYCB} & \multicolumn{4}{c}{HO3D}  \\
\cmidrule(lr){2-3} \cmidrule(lr){4-5} \cmidrule(lr){6-9}
& E$_{J}$$\downarrow$ & E$_{V}$$\downarrow$ & E$_{J}$$\downarrow$ & E$_{RV}$$\downarrow$ & E$_{J}$$\downarrow$ & AUC$_{J}$$\uparrow$ & E$_{V}$$\downarrow$ & AUC$_{V}$$\uparrow$ \\
\midrule
*Boukhayma \textit{et al.}~\cite{boukhayma20193d} & 11.0 & 10.9 & - & 27.3 & - & - & - & - \\ 
*Spurr \textit{et al.}~\cite{spurr2020weakly} & 11.3 & - & 7.1 & - & - & - & - & -  \\
Chen \textit{et al.}~\cite{chen2021model} & 11.8 & 11.9 & - & - & 11.5 & 0.769 & 11.1 & 0.778  \\
Ren \textit{et al.}~\cite{ren2022end} & 10.7 & 11.0 & - & - & - & - & - & -  \\
% Tu \textit{et al.}~\cite{tu2023consistent} & - & - & 19.7 & - & - & - & - & - & - & - \\
Jiang \textit{et al.}~\cite{jiang2023probabilistic} & 10.8 & 10.9 & - & - & 10.5 & 0.789 & 10.7 & 0.785 \\
\midrule
\textbf{Ours} & \textbf{8.5} & \textbf{8.9} & \textbf{6.7} & \textbf{22.0} & \textbf{10.0} & \textbf{0.800} & \textbf{9.8} & \textbf{0.804}  \\
\bottomrule
\end{tabular}
\end{center}
\end{table*}

In this section, we showcase the enhanced performance of our approach compared to state-of-the-art (SOTA) methods in the challenging weakly-supervised setting. We summarize the quantitative evaluation on three different datasets: FreiHAND, DexYCB, and HO3D in Table~\ref{tab:sota}. Our method consistently outperforms existing approaches across all three datasets, which include diverse images depicting daily hand poses and hand-object interactions. Specifically, early methods using 2D weak supervision often require training with 3D annotations due to limited constraints on the 3D predictions \cite{boukhayma20193d,spurr2020weakly}. Chen \textit{et al.}~\cite{chen2021model} avoid the dependency on 3D data by employing different statistical regularizations during training, achieving performance comparable to that of methods utilizing 3D data. Ren \textit{et al.}~\cite{ren2022end} further enhance the performance by leveraging feature consistency constraints. However, these methods are confined to heuristic constraints, such as enforcing a mean pose prediction, or partial types of hand knowledge, like hand biomechanics alone. In contrast, we systematically study and exploit generic hand knowledge, resulting in significant performance improvements. Notably, our improvements over these methods are achieved even without utilizing the NLL loss. For instance, on the FreiHAND dataset, our method achieves E$_{J}$ of 9.4mm, reducing the second-best's 10.7mm by 12\%. Further utilization of the NLL leads to a more significant error reduction of 21\%. Additionally, Jiang \textit{et al.}~\cite{jiang2023probabilistic} propose a probabilistic framework to combine model-based and model-free reconstruction models. Despite their incorporation of additional models, our method outperforms them by a large margin. For example, E$_{V}$ is decreased from 10.9mm to 8.9mm on FreiHAND, and from 10.7mm to 9.8mm on HO3D. Particularly, our method achieves the improved performance by effectively utilizing the generic hand knowledge and modeling the input uncertainty.

\section{Discussion}
\label{sec:discussion}

\begin{table}[tb]
\caption{\textbf{Benefits of Utilizing Generic Hand Knowledge When 3D Annotation is Available.} The units of E$_{J}$, E$_{V}$, and E$_{RV}$ are in mm.}
\label{tab:discussion}
    \begin{subtable}[h]{0.48\textwidth}
        \centering 
        \captionsetup{font=small}
        \caption{\textbf{Generalization.} The evaluation is on DexYCB. The models are trained using 2D hand landmark annotation. ``FreiHAND-3D'' exploits 3D annotated images from FreiHAND to regularize the training on DexYCB, while ``Ours'' uses generic hand knowledge as the prior.}
        \begin{tabular}{l cc }
        \toprule
        Prior& E$_{J}$$\downarrow$ & E$_{V}$$\downarrow$ \\
        \midrule
        FreiHAND-3D & 8.3 & 8.5 \\ 
        \midrule
        Ours & 6.7 & 7.1   \\
        \bottomrule
        \end{tabular}
    \end{subtable}
    \hfill
    \begin{subtable}[h]{0.48\textwidth}
        \centering
        \captionsetup{font=small}
        \caption{\textbf{Data Efficiency.} The evaluation is on FreiHAND. The models are trained using 2D hand landmark annotation and incorporating different percentages of 3D annotation during training.}
        \begin{tabular}{l ccc }
        \toprule
        3D Annotation & E$_{J}$$\downarrow$ & E$_{V}$$\downarrow$ & E$_{RV}$$\downarrow$  \\
        \midrule
        100\% & 8.30 & 8.4 & 21.5\\ 
        \midrule
        Ours (0\%) & 8.52 & 8.9 & 18.4 \\
        Ours+10\% & 8.26 & 8.6 & 16.9 \\
        % 100\% & 8.30 & 8.36 & 21.45\\ 
        % \midrule
        % Ours (0\%) & 8.52 & 8.90 & 18.38 \\
        % Ours+10\% & 8.26 & 8.63 & 16.91 \\
        \bottomrule
        \end{tabular}
     \end{subtable}
\end{table}

In Sec.~\ref{sec:experiment}, we validated our method under the challenging weakly-supervised setting. Here, we demonstrate the advantages of leveraging generic hand knowledge even when 3D annotations are available. Table~\ref{tab:discussion}(a) shows that our method's performance compares favorably with that attained by using a data-driven prior extracted from FreiHAND (following~\cite{kanazawa2018end}) and then evaluated on DexYCB. Thus our method's use of generic hand knowledge gives it a significant advantage over data-driven, domain-specific approaches. Furthermore, our method can take advantage of 3D annotations when they are available. As illustrated in Table~\ref{tab:discussion}(b), when not leveraging any 3D annotation, our method performs just slightly worse than the fully-supervised model (``100\%'') on 2 of the 3 metrics, and it performs comparably to the fully-supervised model using only 10\% of the 3D annotations (``Ours+10\%''). The advantages of generic hand knowledge, including its generalizability and its role in improving data efficiency of monocular 3D hand reconstruction models, further demonstrate its significance.

\section{Conclusion}
We comprehensively study generic hand knowledge, including hand biomechanics, functional anatomy, and physics. We effectively encode these foundational insights as differentiable prior losses, enabling the training of 3D hand reconstruction models solely using 2D annotation. Moreover, we explicitly model image uncertainty with a simple yet effective Negative Log-Likelihood (NLL) loss that incorporates the well-captured uncertainty into the training loss function. Our method significantly outperforms existing weakly-supervised methods. On the widely adopted FreiHAND dataset, the improvement is nearly 21\%.

\noindent\textbf{Society Impact.} Our work highlights the importance of integrating hand knowledge and modeling uncertainty to produce reliable predictions, grounded in hand mechanics and with confidence estimates. It can potentially benefit many downstream tasks like synthetic data generation, biomechanics, and robotics. 

\noindent\textbf{Limitations \& Future Work.} Our method focuses on static generic hand knowledge for image-based reconstruction. A natural extension to our work would be to estimate hand dynamics from monocular videos.

\section*{Acknowledgement} 
This work is supported in part by IBM through the IBM-Rensselaer Future
of Computing Research Collaboration.

% \clearpage  % TODO REVIEW/FINAL: This \clearpage needs to be removed from both review and camera-ready versions.

% ---- Bibliography ----
%
% BibTeX users should specify bibliography style 'splncs04'.
% References will then be sorted and formatted in the correct style.
%
\bibliographystyle{splncs04}
\bibliography{main}
\end{document}